\documentclass[10pt, a4paper]{article}

\usepackage{lrec-coling2024} 

\usepackage{graphicx}
\usepackage{bm}
\usepackage{amsmath, amssymb}
\usepackage{multirow}
\usepackage{booktabs}
\usepackage{longtable}

\title{A Gaze-grounded Visual Question Answering Dataset \\for Clarifying Ambiguous Japanese Questions}

\name{Shun Inadumi${}^{1,2}$, Seiya Kawano${}^{2,1}$, Akishige Yuguchi${}^{3,2}$\\ {\bf \large Yasutomo Kawanishi${}^{2,1}$}, {\bf \large Koichiro Yoshino${}^{2,1}$}} 

\address{   ${}^{1}$ Nara Institute of Science and Technology, Nara, Japan\\
            ${}^{2}$ Guardian Robot Project, RIKEN, Kyoto, Japan 
            ${}^{3}$ Tokyo University of Science, Tokyo, Japan\\
         inazumi.shun.in6@naist.ac.jp, akishige.yuguchi@rs.tus.ac.jp\\
\{seiya.kawano,yasutomo.kawanishi,koichiro.yoshino\}@riken.jp}

\abstract{
Situated conversations, which refer to visual information as visual question answering (VQA), often contain ambiguities caused by reliance on directive information.
This problem is exacerbated because some languages, such as Japanese, often omit subjective or objective terms.
Such ambiguities in questions are often clarified by the contexts in conversational situations, such as joint attention with a user or user gaze information.
In this study, we propose the Gaze-grounded VQA dataset (GazeVQA) that clarifies ambiguous questions using gaze information by focusing on a clarification process complemented by gaze information.
We also propose a method that utilizes gaze target estimation results to improve the accuracy of GazeVQA tasks.
Our experimental results showed that the proposed method improved the performance in some cases of a VQA system on GazeVQA and identified some typical problems of GazeVQA tasks that need to be improved.
\\ 
\newline  }

\begin{document}

\maketitleabstract

\section{Introduction}

The development of interactive systems that can collaborate with humans by taking into account real-world information is one ultimate goal of vision-and-language research.
Such systems should understand the given visual information to respond users based on their results.
Visual question answering (VQA)~\cite{original_VQA, balanced_vqa_v2, C18-1163} and visual dialog~\cite{visdial, agarwal-etal-2020-history} have been proposed to achieve this goal.

VQA tasks generally assume a situation where the intention of questions is clear in visual contexts, and systems can uniquely answer them.
However, in actual interaction with humans, human utterances contain various ambiguities~\cite{survey_language_and_robotics, 4399120}.
A typical problem is exemplified by directives. 
To properly understand questions that contain directives, the directive's destination must be grounded in the real world.
For example, ''Could you pass it to me?'' might have numerous interpretations because of the directive, ``it.''
Some languages, such as Japanese, feature the ellipsis of such topical terms as subject and object in addition to the occurrence of indicative words~\cite{seki-etal-2002-probabilistic, sasano-etal-2008-fully}.
Referring to real-world information is one key idea to resolve the ambiguity caused by directives and ellipses.
For example, speaker's gaze~\cite{EMERY2000581}, speaker's pointing~\cite{nakamura2023ICCV}, and joint attention~\cite{rocca2018CogSci} are important cues for clarifying the target of ellipses and directives.

\begin{figure}[t]
    \centering
    \includegraphics[keepaspectratio, scale=0.20]{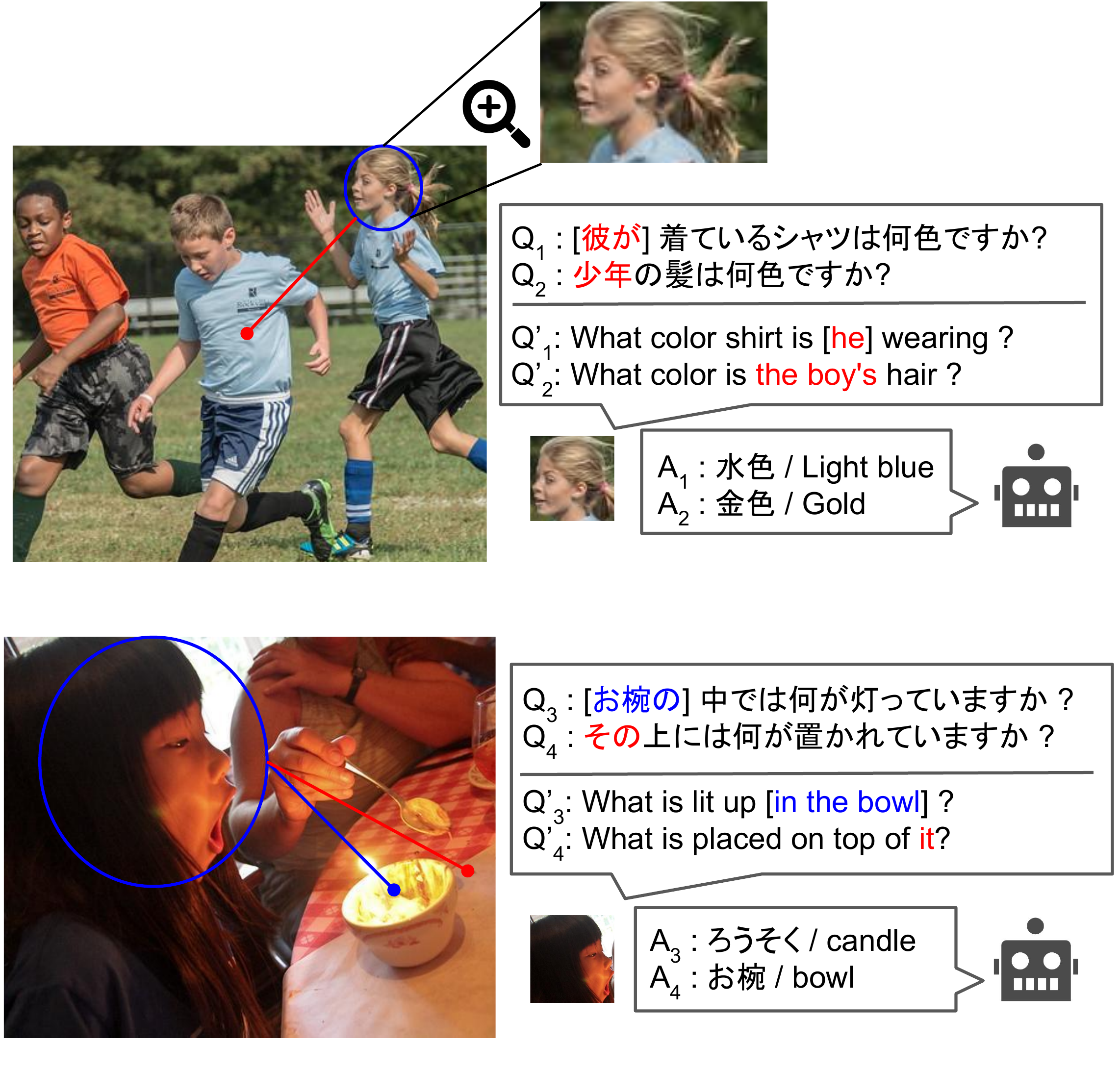}
    \caption{Examples of questions and answers for GazeVQA proposed in this research:  Square brackets denote omitted gaze target names. Multiple target points are assigned that correspond to source points.}
    \label{fig:1}
\end{figure}

In this research, we address the problem of ambiguity in human questions, especially when it refers to gaze information.
We propose a Gaze-grounded VQA dataset (GazeVQA) that includes Japanese questions and gaze information\footnote{Our dataset is publicly available at \url{https://github.com/riken-grp/GazeVQA}.}.
GazeVQA assumes a situation where a speaker in an image asks to a system an ambiguous question that may contain directives or abbreviations, and a system answers it taking into account the speaker's gaze information.
For example, since there are two boys in the upper image in Figure~\ref{fig:1}, the following question from the girl is ambiguous: "What color is the boy's hair?" However, if the system knows the girl's gaze information, it can clarify the ambiguity of the question and answer the question. 
We collected questions and answers by focusing on speaker's gaze targets by crowdsourcing on the MS-COCO (COCO) subset derived object recognition image dataset~\cite{10.1007/978-3-319-10602-1_48} in Gazefollow~\cite{nips15_recasens}.
We collected questions that are difficult to answer without information about speaker's gaze targets and required that the workers not mention the names of the gaze target objects when they created their questions.
As a result, GazeVQA contains 17,276 QA pairs for 10,760 images, of which 1,680 were used as the test-set.
To ensure diverse answers, we assigned ten answers to each question in the GazeVQA test-set.
Our primary contribution is the construction of GazeVQA.

In addition, we propose a model that accurately answers ambiguous questions using gaze information.
Existing vision-and-language models can take a target image and a question about it as input and generate an answer ~\cite{pmlr-v139-cho21a, mokady2021clipcap}.
In this research, we investigate whether models can improve QA accuracy using areas highlighted by gaze information.
A study on segmentation using text and images as prompts~\cite{lueddecke22_cvpr} is related to our idea.
Inspired by this work, we added an adapter consisting of linear layers~\cite{Dumoulin2018_FiLM} to a baseline~\cite{mokady2021clipcap} consisting of a pre-trained image encoder~\cite{pmlr-v139-radford21a} and a text decoder~\cite{radford2019language}.
We proposed a method for integrating a regions of interest (RoI) that represent gaze targets into the whole image with adapters.
We used an existing gaze target estimation model for the estimated a RoI~\cite{Chong_2020_CVPR}.

In experiments, we pre-trained a baseline and the proposed models with a Japanese caption dataset~\cite{yoshikawa-etal-2017-stair} and a Japanese VQA dataset~\cite{C18-1163} and fine-tuned them on our GazeVQA dataset.
In the experimental conditions, we compared the results with and without gaze information in the adapter (ground-truth RoI and estimated RoI).
Our experimental results found that using gaze information improved the GazeVQA’s performance in some cases.
Our second contribution is a proposal of a model that integrates gaze information.

\section{Related Work}
\subsection{Visual Question Answering with Contextual Information}
Visual Question Answering (VQA) is a task where the system derives answers to questions about images~\cite{original_VQA, balanced_vqa_v2, C18-1163}.
Since this study targets questions that are ambiguous without gaze information, we intentionally collected questions that did not include the names of gaze objects.

Previous works proposed VQA datasets that contain a variety of contextual information in addition to images and questions.
A visual dialog provides accurate answers to ambiguous questions that arise during dialogues ~\cite{visdial, agarwal-etal-2020-history}.
To answer questions, the previous dialog history is used as a supplement.
VQA-HAT~\cite{das-etal-2016-human} and VQA-MHUG~\cite{sood-etal-2021-vqa} improved the accuracy of VQA tasks using a saliency map.
By incorporating the answerer's subjective gaze information generated while solving the question with the VQA model, both works grounded fine-grained visual and linguistic representations.
Point and Ask~\cite{mani2020point} employed pointing information to answer ambiguous questions that contain directives.
They used paintings to ground directives in questions and the objects in images.
In our study, we exploit gaze information to answer ambiguous questions.

There are two differences between these previous works and our work on the GazeVQA.
First, we use the questioner's gaze information from the images as additional contextual information.
Second, GazeVQA questions contain not only directives but also Japanese subject and object ellipsis.

Some research that uses gaze information is based on the gazing information of a user looking at an image~\cite{ilaslan-etal-2023-gazevqa}.
In this research, we assume applications such as robots that need to understand the situation from a third person view by extracting the questioner from the image.

\begin{figure*}[t]
    \centering
    \includegraphics[keepaspectratio, scale=0.40]{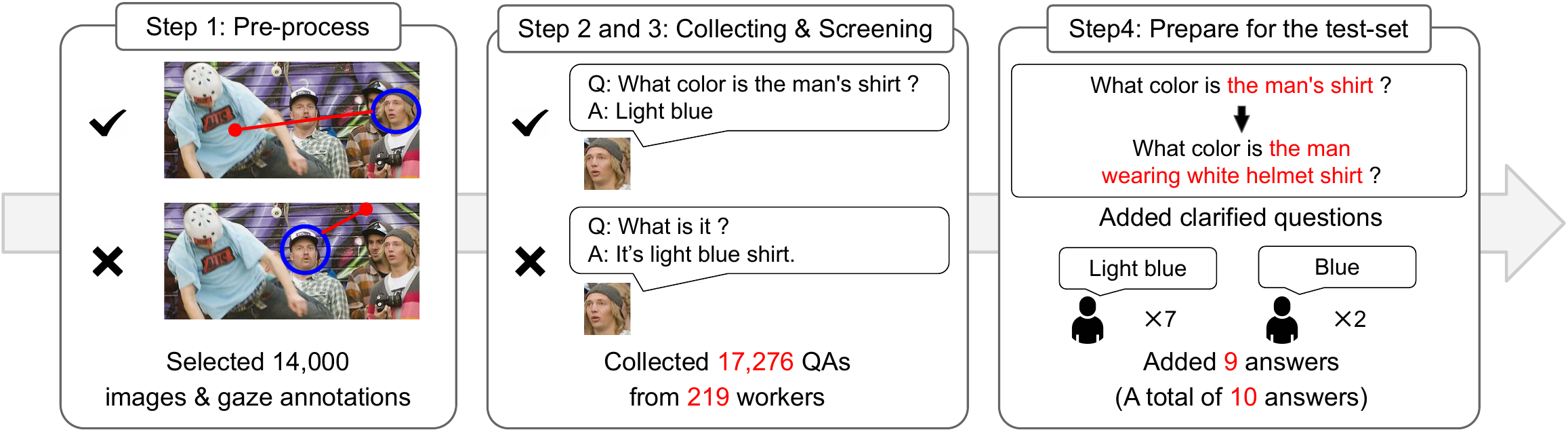}
    \caption{Data collection process of our Gaze-grounded VQA dataset}
    \label{fig:2_1}
\end{figure*}

\subsection{Gaze Target Estimation}
Gaze target estimation predicts a person’s gaze target from a head image.
Gazefollow~\cite{nips15_recasens} is a gaze target estimation dataset that is annotated with the sources and target points as gaze information.
Gazefollow covers people collected from various image datasets, including COCO~\cite{10.1007/978-3-319-10602-1_48}.
Another work maps gaze information to objects for images taken in retail environments~\cite{tomas2021goo}.
In this research, we constructed GazeVQA based on Gazefollow since no special environments are assumed.
Here the gaze destinations in Gazefollow do not necessarily refer to gaze targets; nor do they specify their names.
Therefore, we collected questions and answers about gaze objects based on the object annotations in the COCO subset of Gazefollow.
For the actual gaze target estimation, we used the head image of the person associated with the gaze source~\cite{Chong_2018_ECCV}.

\begin{figure*}[t]
    \centering
    \includegraphics[keepaspectratio, scale=0.47]{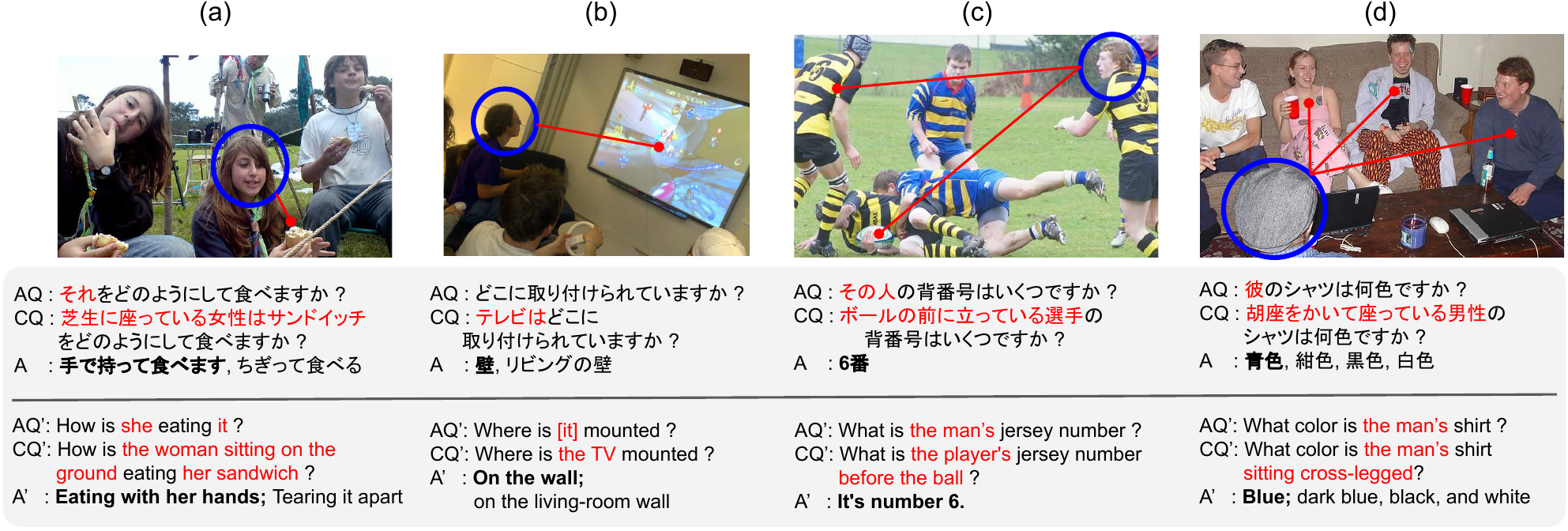}
    \caption{Examples of GazeVQA test-set: AQ and answers in bold denote ambiguous questions and answers obtained through Step 3. CQ denotes questions clarified by annotator's work. The original questions and answers are given in Japanese. We put English translation in the bottom. Words denoted by square brackets are supplements in translations; the terms are omitted in the original Japanese questions.}
    \label{fig:3_1}
\end{figure*}
\section{Gaze-grounded VQA Dataset}
In this section, we describe our proposed Gaze-grounded VQA dataset (GazeVQA).
As in the case of VQA, GazeVQA’s task is to answer questions about the given image.
However, the questions in GazeVQA contain ambiguities and require consideration of the gaze information from the person in the image.
We first describe GazeVQA’s task settings and then explain the data collection process.
We also describe GazeVQA’s statistics.

\subsection{Task Setting}
We consider a case where questions without contextual information are given by a speaker from the system's first-person view.
Models must clarify any ambiguity using the estimated region of interest (RoI) that represents gaze target.
The main task is defined as follows:

\textbf{GazeVQA task:}
Given question $\bm{q}$, corresponding image $I$, and a RoI $I_s$, the task outputs answer $\bm{y}$.

Our data also contain the ground-truth RoI, which is the COCO bounding box; however, we assume that this RoI is not given in a real task. 
In this case, the system also needs to solve the following gaze target estimation task, which is defined as follows to obtain $I_s$:

\textbf{Gaze target estimation task:}
Given image $I$ and speaker's head image $I_h$, the task outputs $I_s$.

\subsection{Data Collection}
Figure~\ref{fig:2_1} shows the process of constructing GazeVQA.
We collected questions and answers for images by crowdsourcing\footnote{\url{https://crowdworks.jp/}}. We used images in the COCO subset of Gazefollow to acquire gold labels of the gaze sources and destinations.
The specific procedure is described below.

\paragraph{Step 1: Selection of images and gaze information:}
We selected 14,000 pairs of image and gaze information and
excluded the following cases: those in which the gaze destinations do not point to objects and those in which the gaze destinations point outside of the image. We used COCO's object segmentation for judgments.
If the gaze destinations do not point to object segmentation, this gaze information is removed.

\paragraph{Step 2: Collecting questions and answers:}
We collected 26,296 questions and answers through crowdsourcing.
Workers wrote questions and answers about gaze targets based on images with gaze information and object labels in COCO.
However, if the gaze targets could not be confirmed due to image blur, we asked them not to create questions and answers for such targets.

The workers were given the following instructions:

\vspace{-0.5\baselineskip}           
{
\setlength{\leftmargini}{9pt}         

\begin{itemize}
	\setlength{\itemsep}{2.5pt}      
	\setlength{\parskip}{0pt}      
	\setlength{\itemindent}{0pt}   
	\setlength{\labelsep}{5pt}     
\item Make the questions at least ten Japanese characters long.
\item Do not include the names of the gaze target objects in the questions.
\item Create questions that can be answered using only the image.

\end{itemize}
}
\vspace{-0.5\baselineskip}           

We designed the first and second instructions to create a variety of ambiguous questions that require gaze information.
We designed the third instruction to exclude from GazeVQA any questions that are ambiguous outside of the image content.
For example, such a question as ``What will he do after this?'' is not covered in this research because it requires some inference.

\paragraph{Step 3: Screening of questions and answers:}
Since the raw crowdsourcing results are noisy, we need to screen them.
In the process of entering questions in Step 2, we placed and used a bonus question: ``Do not enter any text in this field.''
As a post-processing step, we manually checked the unnaturalness of the questions for workers who answered the bonus question.
We excluded the annotation of 27 workers (from the original 246) whose
annotated questions were repetitive or too vague.
We selected 17,276 questions and answers to exclude unnatural questions.
We call this question set ambiguous questions (AQs).

\paragraph{Step 4: Preparation for test-set:}
GazeVQA questions are associated with one or more of the 80 types of gaze objects.
We divided the GazeVQA train/valid/test-set into 13,785/1,811/ 1,860 (0.8 : 0.1 : 0.1).

We expanded the test-set to ensure a variety of answer sets and assigned ten answers to the test-set questions, following a previous work~\cite{original_VQA, balanced_vqa_v2}.
Nine workers created additional answers for each question in the test-set.
Each worker was given only gaze sources and ambiguous questions and answered without gaze destinations and names of the gaze target objects.

We also added a question to clarify each test-set, which contains the names or characteristics of the gaze targets from a single annotator.
The annotator referred to the questions, the answers, and the gaze information.
We called these clarified questions (CQs).

\subsection{Example}
Figure~\ref{fig:3_1} shows a few examples included in the GazeVQA test-set.
These questions suffer from ambiguities due to both directives (Fig.~\ref{fig:3_1} (a)) and ellipsis peculiar to Japanese (Fig.~\ref{fig:3_1} (b)).
The questions are determined as answers based on the content of the questions, even if the gaze targets consist of more than two candidate objects (Fig.~\ref{fig:3_1} (c)).
However, some questions are too vague, where the answers are inconsistent with gaze information (Fig.~\ref{fig:3_1} (d)).

\subsection{Statistics and Analysis}
We compared the statistics of GazeVQA and the Japanese VQA dataset (VQA-ja)~\cite{C18-1163} to highlight the former’s characteristics.

Table~\ref{tab:3_1} shows the statistics of these dataset.
The percentage of unique questions in GazeVQA (46.46\%) exceeds that in VQA-ja (45.21\%), and its average length of questions is also slightly longer.
The percentage of unique answers in GazeVQA (33.87\%) is larger than that in the Japanese VQA (17.10\%), and its average length of answers is also slightly longer.
This is because the GazeVQA questions assumed supplemental information acquired by gaze information in addition to the question itself.

Table~\ref{tab:3_2} shows the typology of question types included in GazeVQA.
The percentage of ``what'' types is 81.85\%, which is about 10\% higher than the percentage of the VQA-ja~\cite{C18-1163}.
GazeVQA includes many questions that ask about the attributes of the gaze target object, such as color and shape, because the gaze target was the question’s subject.
The percentage of ``where'' types about the location of objects and ``how'' types about the number of objects was 12.04\% in total.
GazeVQA also contains other types of questions, including ``when'' types that ask about time and ``who'' types that ask about a person.

\begin{table}[t]
    \centering
    \caption{Statistics on GazeVQA and Japanese VQA (VQA-ja)~\cite{C18-1163}}
        \begin{tabular}{lrr} 
        \toprule
         & GazeVQA & VQA-ja \\
         \cmidrule(lr){0-2}
        Images & 10,760 & 99,208 \\
        Question and answers & 17,276 & 793,664 \\
        Unique questions & 8,628 & 358,844 \\
        Unique answers & 5,853 & 135,743 \\
        \cmidrule(lr){0-2}
        Avg. question length & 15.37 & 14.82 \\
        Avg. answer length & 4.92 & 4.56 \\
        \bottomrule
        \end{tabular}
    \label{tab:3_1}
\end{table}

\begin{table}[t]
    \centering
    \caption{Typology of question types for GazeVQA}
        \begin{tabular}{l|r} 
        \hline
        Types (Keywords in Japanese)& \#Counts\\\hline
        What\quad(\textit{nani}, \textit{dono}, \textit{donna}) & 14,141 \\
        \quad is/are/do/does &  7,215\\
        \quad color &  3,626 \\
        \quad condition &1,240\\
        \quad kind & 903 \\
        \quad shape &  703 \\
        \quad others &  454 \\\hline
        Where\ (\textit{doko}) &  1,085 \\
        How\quad\ (\textit{dore}, \textit{ikutsu}) & 996 \\
        Which\ \ (\textit{dochira}) & 295 \\
        Others\ (\textit{itsu}, \textit{dare}, \textit{naze}) & 875 \\\hline
        \end{tabular}
    \label{tab:3_2}
\end{table}

Table~\ref{tab:3_3} shows the frequency of arguments in the predicate-argument structure\footnote{We calculated this frequency through an integrated Japanese text analyzer~\cite{ueda-etal-2023-kwja}.} in the GazeVQA test-set.
Ambiguous questions in the GazeVQA test-set often result in the ellipsis of nominative and accusative cases, related to subjects and objects in questions, compared with clarified questions.
This result suggests that GazeVQA contains questions in which the nominative and accusative cases are omitted, which is often in Japanese.
For example, Fig.~\ref{tab:3_3}(b) is a typical example of the ellipsis of the nominative case.

\begin{table}[t]
    \centering
    \caption{Frequency of predicate term relationships in test-set of GazeVQA questions: Note that ``nom.'', ``acc.'' and ``dat.'' denote numbers of nominative, accusative and dative cases, AQ and CQ refer to caption of Fig.~\ref{fig:3_1}.}
        \begin{tabular}{c|r|r|r} 
        \hline
        Types & \textit{ga} (nom.) & \textit{wo} (acc.) & \textit{ni} (dat.) \\\hline
        AQ & 2,044 & 1,028 & 440  \\
        CQ & 2,912 & 1,584 & 569 \\\hline
        \end{tabular}
    \label{tab:3_3}
\end{table}

\begin{figure*}[t]
    \centering
    \includegraphics[keepaspectratio, scale=0.65]{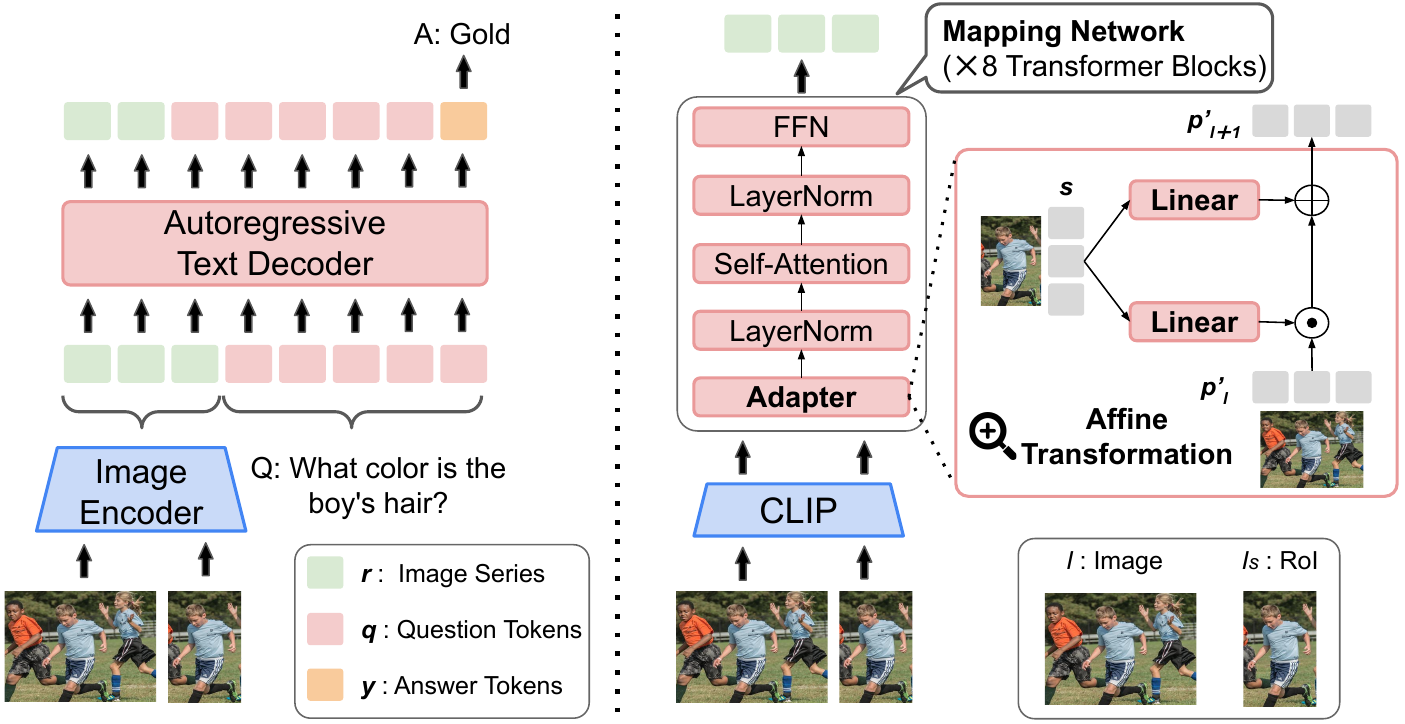}
    \caption{\textbf{Left:} Overview of proposed system \textbf{Right: } Details of Image Encoder architecture}
    \label{fig:4_1}
\end{figure*}

\section{Methodology}
In this section, we first describe ClipCap~\cite{mokady2021clipcap}, which is our baseline for the GazeVQA task, and next describe our proposed model, ``ClipCap + Adapter,'' which adds adapters~\cite{Dumoulin2018_FiLM} to ClipCap.
Finally, we explain the procedure for obtaining the region of interest (RoI) of the gaze targets in the gaze target estimation task.

\subsection{Baseline Model: ClipCap}
ClipCap is a vision-and-language model consisting of an image encoder and a text decoder.
We used ClipCap as the baseline for the GazeVQA task because there is no representative VQA model pre-trained in Japanese.

\paragraph{Image encoder:}
Given a RGB image $I \in \mathbb{R}^{W \times H \times 3}$, the baseline image encoder outputs image series $\bm{r} = \{r_1, \dots, r_n\}$ that can be input to the text decoder.
Here $n$ is the length of the image series, and element $r_i$ in $\bm{r}$ has the same dimensions as the token embedding in question $\bm{q}$.

Given image $I$, CLIP image encoder~\cite{pmlr-v139-radford21a} outputs image series $\bm{p} = \{p_1, \dots, p_n\}$ using a single linear layer $f$:
\begin{align}
\label{form:1}
\{p_1, \dots, p_n\} = f(CLIP(I)).
\end{align}
Given image series $\bm{p}$, the multi-layer transformer blocks~\cite{10.5555/3295222.3295349} the $F$ output $\bm{r}$:
\begin{align}
\label{form:2}
\{r_1, \dots, r_n\} = F(\{p_1, \dots, p_n\}).
\end{align}
We call these transformer blocks a mapping network, following the previous work~\cite{mokady2021clipcap}.

\paragraph{Text decoder:}
Given question tokens $\bm{q} = {q_1, \dots, q_m}$ and image series $\bm{r}$, the autoregressive text decoder generates answer tokens $\bm{y}$.
The following is the input series of the text decoder:
\begin{align}
\label{form:3}
\{r_1, \dots, r_n, [SEP1], q_1, \dots, q_m, [SEP2]\},
\end{align}
where $[SEP1]$ and $[SEP2]$ are ``Question:'' and ``Answer:'' and represent the decoder prompts.

\subsection{Proposed Model: ClipCap + Adapter}
Figure~\ref{fig:4_1} shows the structure of our proposed model.
We added adapters to a mapping network~\cite{Dumoulin2018_FiLM}, inspired by work on object segmentation using text and objects as queries~\cite{lueddecke22_cvpr}.
Adapters merge image $I$ and RoI $I_s$, and the mapping network outputs an image series that takes into account a gaze target.
Each mapping network's transformer block has an adapter (Fig.~\ref{fig:4_1}, right).
The CLIP image encoder constructs two image series: one for image $\bm{p}$ and another for the RoI of the gaze target $\bm{s} = \{s_1, \dots, s_n\}$ from $I$ and $I_s$, similar to the baseline image encoder.
Given $\bm{p}$ and $\bm{s}$, the adapter computes the element-wise affine transformation and outputs a mixture of features $\bm{p'}$ from $I$ and $I_s$:
\begin{align}
\label{form:4}
\bm{p'}_{l+1} = g(\bm{s}) \odot \bm{p'}_{l} \oplus h(\bm{s}),
\end{align}
where $g$ and $h$ denote a linear layer and $\bm{p'}_{l}$ denotes the input of the transformer block of the mapping network in the $l$-th layer.
Note that the input of the first transformer block is $\bm{p'}_{l} = \bm{p}$.

\subsection{Process of Gaze Target Estimation}
We obtain RoI $I_s$, which is an input to the adapter, from a head image of gaze source $I_h$.
Given image $I$ and head image $I_h$, the gaze target estimation model~\cite{Chong_2020_CVPR} outputs a gaze heatmap $H$.
We binarize a threshold value of 0 for $H$ and obtain $I_s$, which is a bounding box corresponding to the gaze target~\cite{Ardizzone2013_saliency}. 
We consider $I$ to be $I_s$ since it is difficult to get $I_s$ from $H$ if every element of $H$ is 0.

\section{Experiments}
\subsection{Experimental Setup}
\label{sec:setup}
\paragraph{Dataset:}
We used 123,287 images and 616,435 captions from the Japanese image caption dataset~\cite{yoshikawa-etal-2017-stair} (STAIR) and 99,208 images and 793,664 question-answer pairs from the Japanese VQA dataset~\cite{C18-1163} (VQA-ja) as pre-training for the models.
We fine-tuned them using the GazeVQA train-set.

\paragraph{Implementation details:}
We used a ResNet-based $RN\times4$~\cite{pmlr-v97-tan19a} as the CLIP image encoder and processed images $I$ and regions of interest $I_s$ in a manner that resembles CLIP normalization\footnote{\url{https://github.com/openai/CLIP}}.
The input of the CLIP image encoder is a resized image with 224 dimensions (height and width); the output is a 640-dimensional vector.
We composed a mapping network of eight layers of transformer blocks and set length $n$ of the image series ($\bm{p}$, $\bm{s}$, and $\bm{r}$) to 10.
We used GPT-2 as our text decoder~\cite{pmlr-v139-radford21a}, which was pre-trained on a Japanese corpus\footnote{\url{https://huggingface.co/rinna/japanese-gpt2-medium}}.

For a batch size of 32, we trained 10 epochs for STAIR, VQA-ja, and GazeVQA.
The optimizer was AdamW~\cite{loshchilov2018decoupled}, with a learning rate of 2e-5 in pre-training and 1e-4 in fine-tuning.
We used a beam search with beam width of 10 for the GazeVQA evaluation.

\paragraph{Training target:}
We next describe the results of training the parameters of the mapping network and the text decoder due to the limited data available in Japanese.
Our model has about 426M training parameters: 410M baseline training parameters and 16M adapter parameters.
We also report the results of training the mapping network or only the adapters with GazeVQA to explicitly update the adapter weights.
There are 74M baseline training parameters, and our model has 90M training parameters only when the mapping network is trained.

\paragraph{Evaluation metrics:}
We evaluated the model with VQA score $Acc$ that takes into account the diversity of the answers in the VQA task~\cite{original_VQA, balanced_vqa_v2}.
We also evaluated the model using a BERT score, $Bs$~\cite{Zhang*2020BERTScore:}, which takes into account the variability of the responses.
We used a multilingual BERT sentence vector for our evaluation and calculated the 
similarity of the vectors between the predicted answer and each element in the gold answer set\footnote{\url{https://huggingface.co/bert-base-multilingual-cased}}. 
$Bs$ is the arithmetic mean of all these similarities.

\subsection{Quantitative Evaluation}
Table~\ref{tab:5_1} shows the evaluation results of the proposed model and the baseline.
Table~\ref{tab:5_2} shows the ablation study results for the baseline inputs.
Here all the scores ($Acc$ and $Bs$) are the averages of five training and evaluation iterations of the GazeVQA.
We denote the image as $I$, the RoI obtained from the gaze target estimation as $I_s$, and the gold RoI as $GT$, which is a COCO bounding box associated with the question, with respect to the model inputs.

\begin{table}[t]
    \centering
    \caption{Evaluation results of baseline and proposed models with GazeVQA test-set: $|\theta|$ is number of trainable parameters for each model. }
    \begin{tabular}{lrrr}
    \toprule
     Models  & $|\theta|$ & Acc & Bs \\
    \cmidrule(lr){0-3} 
    \multicolumn{4}{c}{Fine-tuned Text Decoder \& Mapping Network}
    \\
    \cmidrule(lr){0-3}
    ClipCap & 410 & \bf{36.80} & \bf{81.75} \\
    ClipCap + Adapter $(I)$ & 426 & 34.78 & 81.39 \\
    ClipCap + Adapter $(I_s)$ & 426 & 34.15 & 81.28 \\
    ClipCap + Adapter $(GT)$ & 426 & 34.72 & 81.33 \\
    \cmidrule(lr){0-3} 
    \multicolumn{4}{c}{Fine-tuned Mapping Network} \\
    \cmidrule(lr){0-3}
    ClipCap & 74 & 35.83 & 81.21 \\
    ClipCap + Adapter $(I)$ & 90 & \bf{38.45} & \bf{81.74} \\
    ClipCap + Adapter $(I_s)$ & 90 & 38.11 & 81.71 \\
    ClipCap + Adapter $(GT)$ & 90 & 38.01 & 81.70 \\
    \cmidrule(lr){0-3}
    \multicolumn{4}{c}{Fine-tuned Adapter Only} \\
    \cmidrule(lr){0-3}
    ClipCap + Adapter $(I)$ & 16 & 40.06 & 81.91 \\
    ClipCap + Adapter $(I_s)$ & 16 & 39.03 & 81.92 \\
    ClipCap + Adapter $(GT)$ & 16 & \bf{40.09} & \bf{82.01} \\
    \bottomrule
    \end{tabular}
    \label{tab:5_1}
\end{table}

\paragraph{Our model vs. baseline:}
We compared our proposed model (ClipCap + Adapter $(I_s)$) with the baseline (ClipCap) with the RoI $I_s$ input to the adapter.
Appendix~\ref{sec:eval_q_types} shows detailed evaluation results for each question type described according to the classification in Table~\ref{tab:3_2}.

As shown in Table~\ref{tab:5_1}, our model underperformed the baseline when the mapping network and the text decoder are trained with GazeVQA.
However, it outperformed the baseline performance when only the mapping network and the adapters were trained with GazeVQA.
In particular, the VQA score of our model trained only with adapters is 39.03, which is about four points higher than the baseline trained with the mapping network and the text decoder.
Our model can generate accurate answers to ambiguous questions with about 16M parameter updates, compared to the baseline, which requires a full tuning both text decoder and a mapping network.

\paragraph{Factors contributing to GazeVQA task accuracy:}
We compared our proposed model with a baseline trained only on the mapping network.
As shown in Table~\ref{tab:5_1}, our model with image $I$ as input to the adapter (ClipCap+ Adapter (I)) outperformed the baseline, and there is no difference in our model with RoI $I_s$ and $GT$ as input: ClipCap+ Adapter ($I_s$) and ClipCap+ Adapter ($GT$).
This result suggests that the increase in training parameters due to the addition of adapters is one reason for the improved accuracy of the GazeVQA task.
This result also suggests that using RoI $I_s$, which is a model for gaze target estimation, may reduce the accuracy when the estimation is incorrect.
Our qualitative evaluation in Section~\ref{sec:qualitative} discusses these results.

\paragraph{Characteristics of our dataset:}
We identified the elements needed to resolve ambiguous questions in GazeVQA through an ablation study on the baseline.
As shown in Table~\ref{tab:5_2}, the performance of the baseline, which excludes question tokens or image series from the input, is significantly worsened.
Models need to jointly understand the images/questions to solve GazeVQA tasks.

The performance of the baseline with regions of interest ($I_s$ and $GT$) as input to the image encoder falls below the baseline with image $I$ as input.
This result suggests that keeping some information outside the gaze targets, rather than completely removing such information, improves the accuracy of the GazeVQA task.

\begin{table}[t]
    \centering
    \caption{Ablation evaluation results of baseline: $I_s$ and $GT$ denote that baseline only uses a limited region of image $I$ pointed by their bounding boxes.}
    \begin{tabular}{lrr}
    \toprule
     Models & Acc & Bs \\
     \cmidrule(lr){0-2}
     ClipCap & 36.80 & 81.75 \\
     \quad w/o image series & 16.10 & 78.48 \\
     \quad w/o question tokens & 3.66 & 65.93 \\
     ClipCap $(I_s)$ & 34.53 & 81.28 \\
     ClipCap $(GT)$ & 34.27 & 81.26 \\
     \bottomrule
    \end{tabular}
    \label{tab:5_2}
\end{table}

\begin{figure*}[t]
    \centering
    \includegraphics[keepaspectratio, scale=0.49]{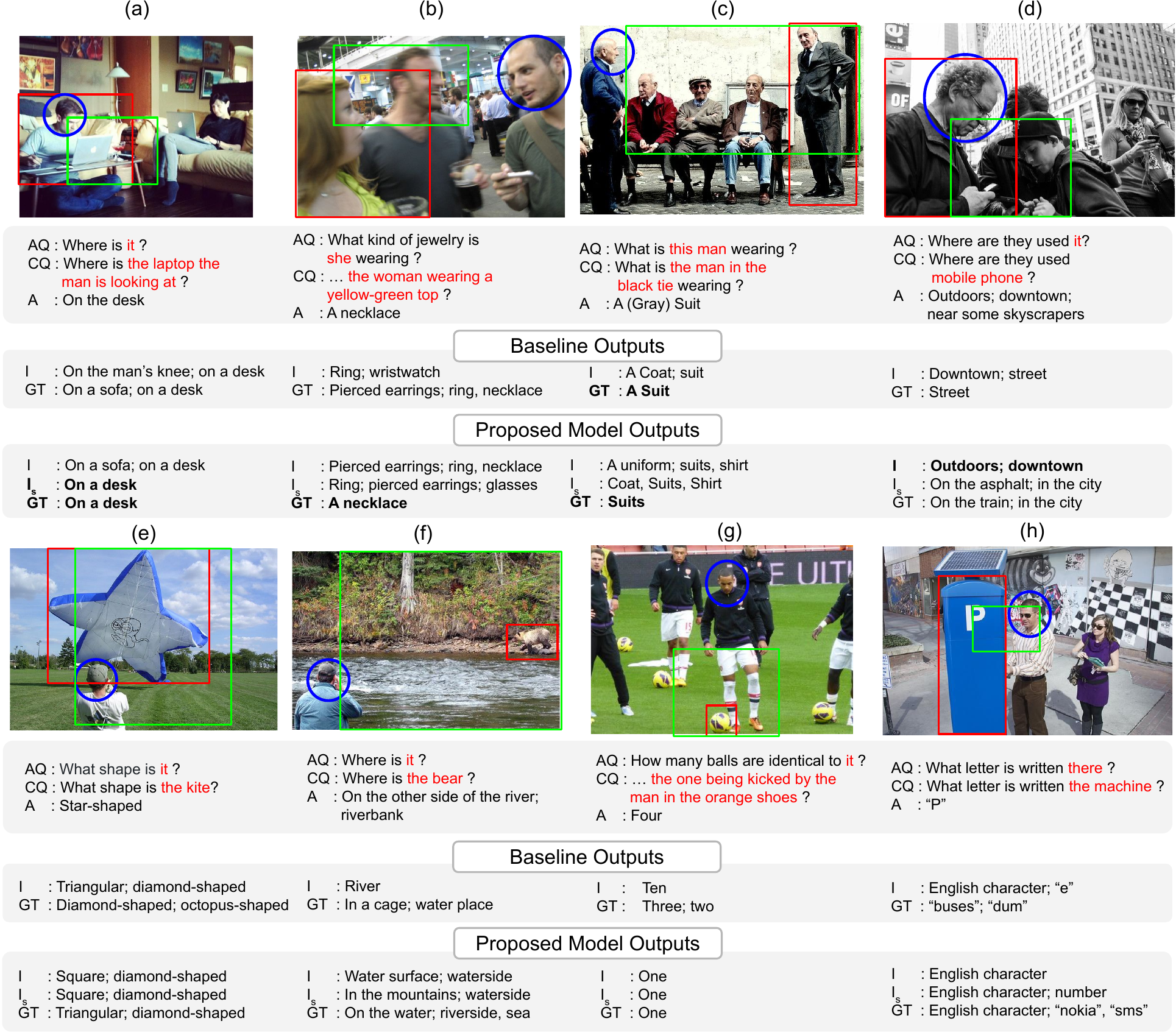}
    \caption{Outputs for baseline and proposed models: AQ, CQ, and A are respectively ambiguous questions, clarified questions, and examples of correct answers. Bolded results are models that scored best among five attempts. $GT$ and $I_s$ are denoted by red and green boxes.}
    \label{fig:5_1}
\end{figure*}

\subsection{Qualitative Evaluation}
\label{sec:qualitative}
Figure~\ref{fig:5_1} shows examples of the actual outputs for the baseline and our models.
We examined the impact of the differences in the inputs to the adapters on the results of our proposed model.
First, our model with the RoI $GT$ input to the adapter tended to provide unique answers to ambiguous questions about the attributes of the gaze targets, such as the object’s shape and name.
As shown in Figures~\ref{fig:5_1} (a) and (b), this tendency is more pronounced when the RoI contains visual features that contribute to providing an accurate answer.
In other words, our model outputs inconsistent answers when the gaze target estimation model cannot narrow down the objects at the gaze target (Fig.~\ref{fig:5_1} (c)).
Finally, our model with image $I$ input to the adapter tends to give accurate answers to questions that require an understanding of the image (Fig.~\ref{fig:5_1} (d)).

\section{Discussion and Limitations}
We proposed GazeVQA to achieve a system that can understand the ambiguities in human utterances using a speaker's gaze information.
The visual features contained in GazeVQA were a single image, and the gaze destinations and sources were within its frame.
However, since the visual features captured by an actual system, such as a robot, are dynamic, they contain uncertainty.
This situation makes it difficult for the system to recognize speakers and disambiguation cues.
We believe we should fully use gaze information and such modalities as pointing~\cite{nakamura2023ICCV} and the dialog context before utterances~\cite{visdial, yu-etal-2019-see} to account for visual uncertainty.

GazeVQA was designed for Japanese questions, and the availability of Japanese vision-and-language data is limited.
For this reason, our study investigated a good training efficiency baseline~\cite{mokady2021clipcap} and method~\cite{Dumoulin2018_FiLM}. 
However, none of the models used in this research accurately answered questions about the shape of special objects, positional relationships, number of objects, or character comprehension (Fig.~\ref{fig:5_1} (e)-(h)).
A system needs to understand the gaze information to identify what is the object indicated by directives or ellipsis, but the question requires information from other areas in the image; as in the case of  Figure~\ref{fig:5_1} (g).
We believe a model structure must be used that allows for a fine-grained understanding of the correspondence between vision-and-language~\cite{pmlr-v139-cho21a,openai2023gpt4} to alleviate this problem.
Appendix~\ref{sec:other_models} shows evaluation results of how these models can handle GazeVQA ambiguous questions and clarified questions.


\section{Conclusion}
We introduced a Gaze-grounded VQA dataset (GazeVQA) to address the problem of ambiguities in human utterances in real world.
Answering GazeVQA questions is challenging without the speaker's gaze information and contains ambiguities about directives and ellipsis peculiar to Japanese.
Furthermore, we proposed a model that integrates the region of interest of the gaze target as gaze information in addition to images and questions.
Quantitative results show that our model improves the performance over a baseline on the GazeVQA task.
Qualitative results show that our model provides accurate answers to ambiguous questions about the attributes of gaze objects through gaze information.
Our future work will address the difficult cases in our study by exploring model architectures and methods for integrating gaze information.

\section{Acknowledgements}
This research was supported by JSPS KAKENHI Grant Number 22H04873 and JST Moonshot R\&D Program Grant Number JPMJMS2236.

\nocite{*}
\section{Bibliographical References}
\label{sec:reference}

\bibliographystyle{lrec-coling2024-natbib}
\bibliography{references}


\appendix
\section{Evaluation by Question Types}
\label{sec:eval_q_types}
\begin{figure*}[t]
    \centering
    \includegraphics[keepaspectratio, scale=0.48]{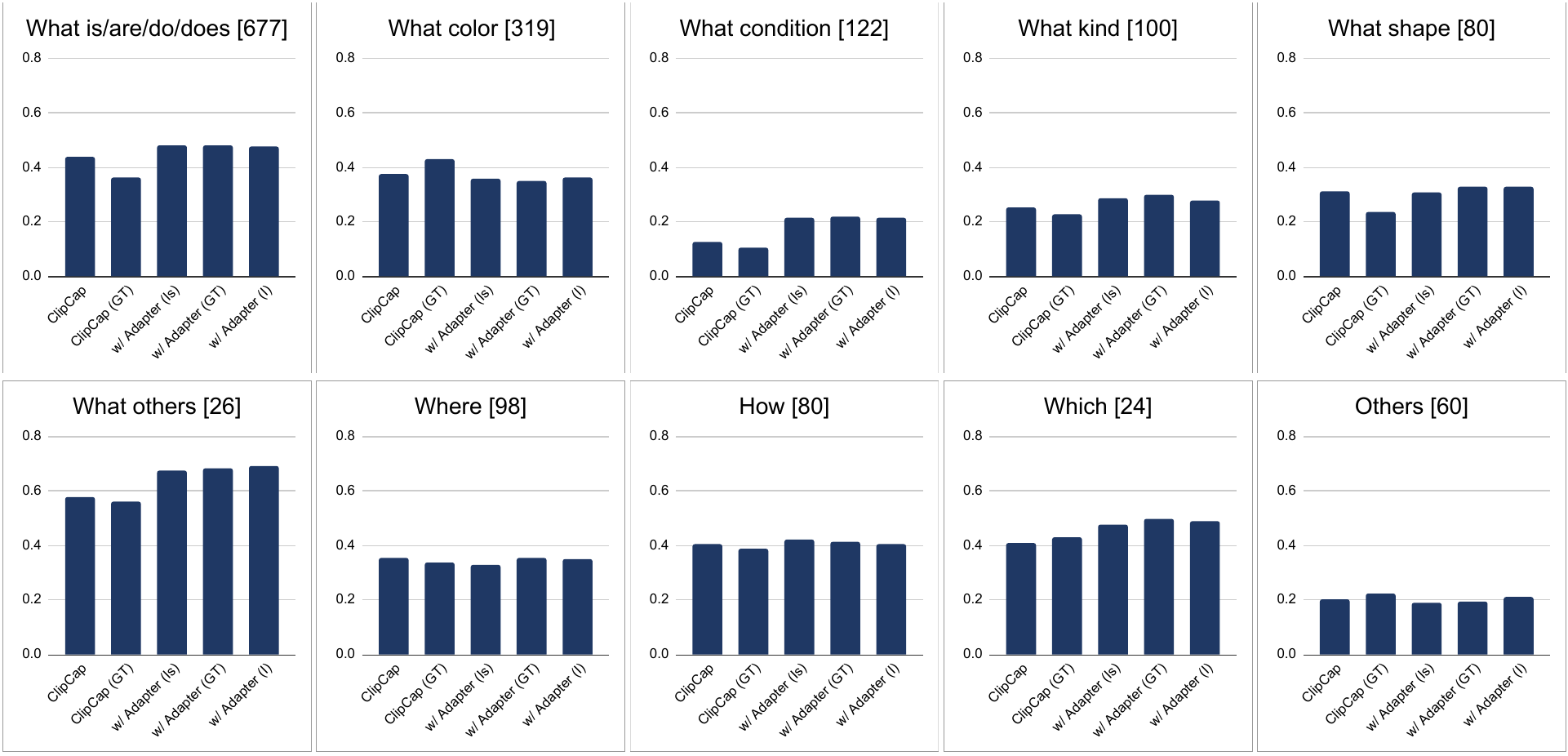}
    \caption{Evaluation results of baseline and proposed model by question types for GazeVQA test-set: Square brackets denote the number of questions.}
    \label{fig:8_1}
\end{figure*}

We compared our proposed model with the baseline based on the typology of question types for GazeVQA shown in Table~\ref{tab:3_2}.
As shown in Figure~\ref{fig:8_1}, our model performs well with ``What is'' questions about object attributes, ``What condition'' questions about the current state of an object, and ``Which'' questions that are multiple choice questions.
The baseline with RoI $GT$ (ClipCap($GT$)) performs well with ``What color'' questions that ask for an object color.

\section{Discussion on Evaluation with Clarified Questions}
\label{sec:other_models}
Figure~\ref{tab:8_1} shows comparative evaluation results of ambiguous questions and clarified questions with the baseline (ClipCap) and modern vision-and-language models: VL-T5~\cite{pmlr-v139-cho21a, sung2022vladapter} and GPT-4V~\cite{openai2023gpt4}.
We used 300 samples from our GazeVQA test-set for evaluation and did not fine-tune any models with GazeVQA train-set.

\subsection{Implementation details}
VL-T5 is a vision-and-language model consisting of an image encoder~\cite{NIPS2015_14bfa6bb} and the text encoder-decoder~\cite{2020t5}. 
We constructed VL-T5 with the CLIP image encoder~\cite{pmlr-v97-tan19a} and the Japanese T5 model \footnote{\url{https://huggingface.co/retrieva-jp/t5-small-short}} \footnote{\url{https://huggingface.co/retrieva-jp/t5-base-short}}, based on the implementation of~\citet{sung2022vladapter}.
We used the same conditions as in Section~\ref{sec:setup} for the VL-T5 training setup.

GPT-4V is a large-scale vision-and-language model trained on large amounts of image-text data.
We evaluated the GazeVQA test-set using GPT-4V in the 3-shot setting; each example was constructed from questions and answers and gaze targets included in the GazeVQA train-set.
Tables~\ref{tab:8_2} and~\ref{tab:8_3} show prompts given to GPT4V for inferring an answer from either an ambiguous question and bounding boxes of gaze targets $GT$ or a clarified question.

\subsection{Results}
Figures~\ref{tab:5_1} and~\ref{tab:8_1} suggest that our model fine-tuned with GazeVQA outperforms GPT4V when ambiguous questions are used.
On the other hand, Figure~\ref{tab:8_1} shows that GPT-4V outperforms other models when clarified questions are used as input instead of ambiguous questions.
These results indicate that large vision-and-language models such as GPT4V are highly capable, but they are not sufficient in situations such as GazeVQA task, where the question contains ambiguity and needs to be supplemented with contextual information.

\onecolumn
\begin{table*}[t]
    \centering
    \caption{Comparison results between ambiguous questions (AQ) and clarified questions (CQ) using vision-and-language models}
    \begin{tabular}{lrrr}
    \toprule
     Models & Types & Acc & Bs \\
     \cmidrule(lr){0-3}
     ClipCap & AQ & 21.55 & 78.19 \\
     ClipCap & CQ & \textbf{25.11} & \textbf{79.41} \\
     \cmidrule(lr){0-3}
     VL-T5 {\scriptsize small} & AQ & \textbf{32.66} & 80.20 \\
     VL-T5 {\scriptsize small} & CQ & 31.66 & \textbf{80.27} \\
     \cmidrule(lr){0-3}
     VL-T5 {\scriptsize base}  & AQ & 32.33 & 80.16 \\
     VL-T5 {\scriptsize base}  & CQ & \textbf{34.11} & \textbf{80.77} \\
     \cmidrule(lr){0-3}
     GPT-4V {\scriptsize 3-shot} & AQ & 34.11 & 79.99 \\
     GPT-4V {\scriptsize 3-shot} & CQ & \textbf{39.33} & \textbf{80.17} \\
     \bottomrule
    \end{tabular}
    \label{tab:8_1}
\end{table*}

\begin{table*}[t]
    \centering
    \caption{Prompts used to evaluate GazeVQA ambiguous questions: gaze targets $GT$ are denoted by red boxes.}
    \begin{tabular}{p{0.95\linewidth}}
    \toprule
    \textbf{Instruction} \\
    \texttt{Instruction: Given an ambiguous Japanese question that includes ellipsis or directives, an image, and bounding boxes (format:[$x_1$,$y_1$,$w$,$h$]), you answer the question in Japanese. Note1: Each question is answerable if you consider the bounding boxes corresponding to the ellipsis or directives of the question. Note2: Each answer will end with a noun.} \\
    \cmidrule(lr){0-0}
    \textbf{Visual input examples} \\
    \begin{minipage}[b]{0.31\linewidth}
        \centering
        \scalebox{0.30}{\includegraphics{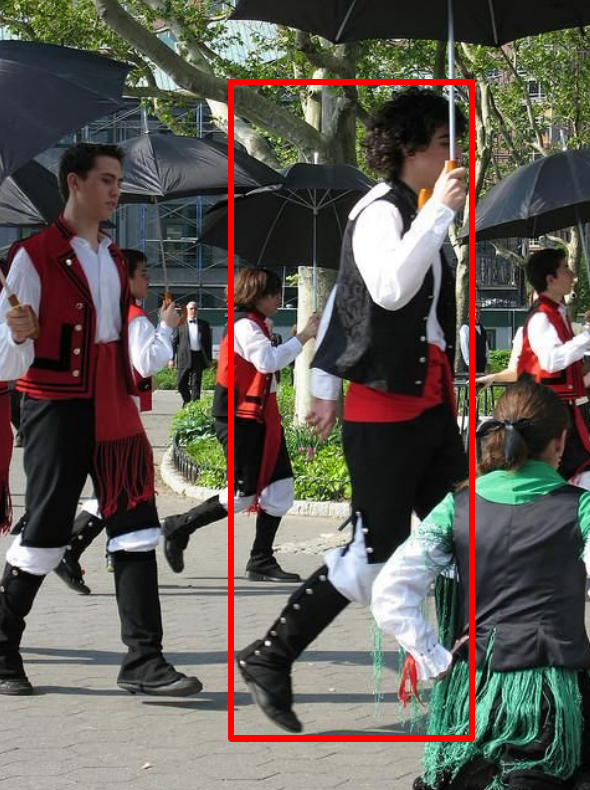}}
        \texttt{\{Example1\}}
    \end{minipage}
    \begin{minipage}[b]{0.31\linewidth}
        \centering
        \scalebox{0.35}{\includegraphics{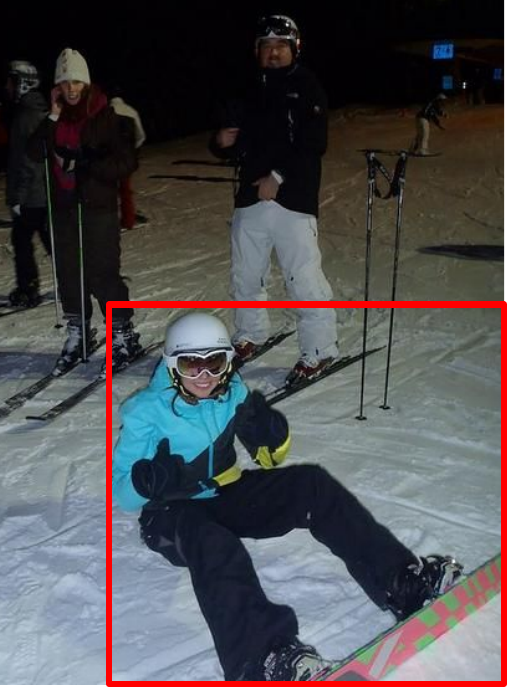}}
        \texttt{\{Example2\}}
    \end{minipage}
    \begin{minipage}[b]{0.31\linewidth}
        \centering
        \scalebox{0.36}{\includegraphics{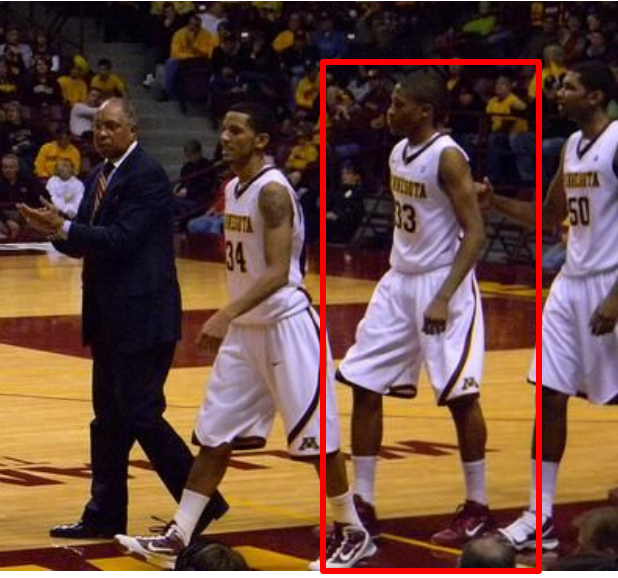}}
        \texttt{\{Example3\}}
    \end{minipage}\\
    \cmidrule(lr){0-0}
    \textbf{Text input example1} \\
    \texttt{The question is: What color is he wearing?}\\
    \texttt{The image is: \{Example1\}}\\
    \texttt{The bounding boxes are: [194.16,69.03,194.15,524.95]} \\
    \texttt{Answer the question with a single phrase in Japanese: Black} \\
    \cmidrule(lr){0-0}
    \textbf{Text input example2} \\
    \texttt{The question is: What is she wearing on her head?}\\
    \texttt{The image is: \{Example2\}}\\
    \texttt{The bounding boxes are: [158.2,339.42,291.96,293.39]} \\
    \texttt{Answer the question with a single phrase in Japanese: Helmet} \\
    \cmidrule(lr){0-0}
    \textbf{Text input example3} \\
    \texttt{The question is: What is his number?}\\
    \texttt{The image is: \{Example3\}}\\
    \texttt{The bounding boxes are: [440.31,156.82,117.88,310.4]} \\
    \texttt{Answer the question with a single phrase in Japanese: Number 33} \\
     \bottomrule
    \end{tabular}
    \label{tab:8_2}
\end{table*}

\begin{table*}[t]
    \centering
    \caption{Prompts used to evaluate GazeVQA clarified questions.}
    \begin{tabular}{p{0.95\linewidth}}
    \toprule
    \textbf{Instruction} \\
    \texttt{Instruction: Given a Japanese question and an image, you answer the question in Japanese. Note: Each answer will end with a noun.}\\
    \cmidrule(lr){0-0}
    \textbf{Visual input examples} \\
    \begin{minipage}[b]{0.31\linewidth}
        \centering
        \scalebox{0.30}{\includegraphics{contents/fig/example1_wo_bbox.pdf}}
        \texttt{\{Example1\}}
    \end{minipage}
    \begin{minipage}[b]{0.31\linewidth}
        \centering
        \scalebox{0.35}{\includegraphics{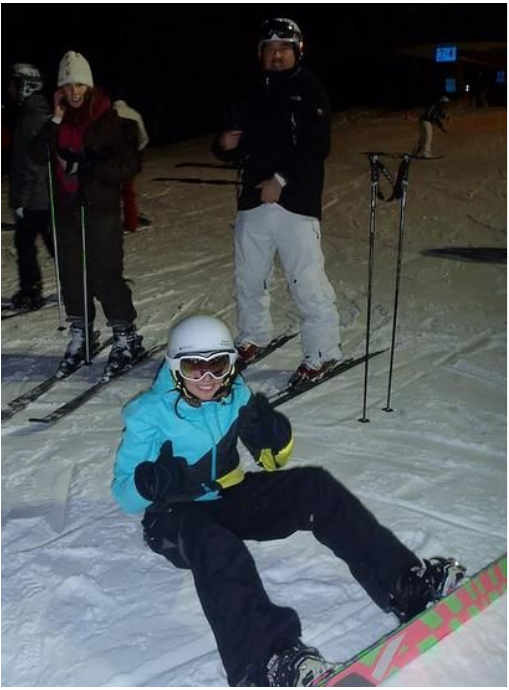}}
        \texttt{\{Example2\}}
    \end{minipage}
    \begin{minipage}[b]{0.31\linewidth}
        \centering
        \scalebox{0.36}{\includegraphics{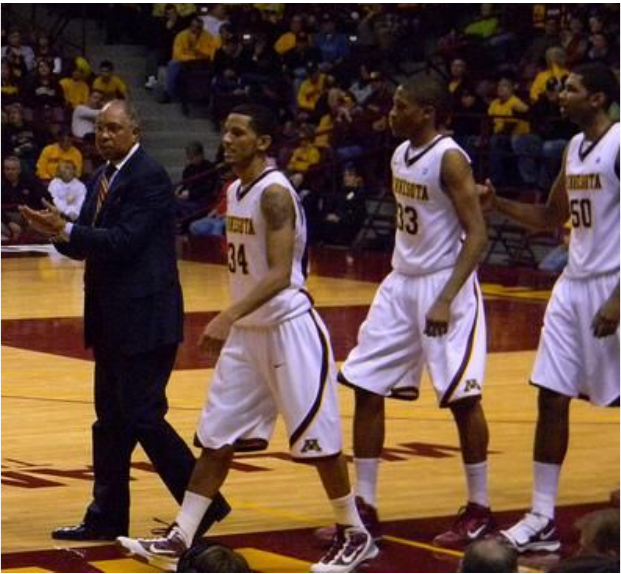}}
        \texttt{\{Example3\}}
    \end{minipage}\\
    \cmidrule(lr){0-0}
    \textbf{Text input example1} \\
    \texttt{The question is: What color is the man on the right with the black umbrella wearing?}\\
    \texttt{The image is: \{Example1\}}\\
    \texttt{Answer the question with a single phrase in Japanese: Black} \\
    \cmidrule(lr){0-0}
    \textbf{Text input example2} \\
    \texttt{The question is: What is the woman in the blue jacket wearing on her head?}\\
    \texttt{The image is: \{Example2\}}\\
    \texttt{Answer the question with a single phrase in Japanese: Helmet}\\
    \cmidrule(lr){0-0}
    \textbf{Text input example3} \\
    \texttt{The question is: What is his number of the second man from the right?}\\
    \texttt{The image is: \{Example3\}}\\
    \texttt{Answer the question with a single phrase in Japanese: Number 33} \\
    \bottomrule
    \end{tabular}
    \label{tab:8_3}
\end{table*}

\end{document}